\documentclass{article}

\PassOptionsToPackage{round}{natbib}

\usepackage[preprint]{neurips_2019}
\usepackage[utf8]{inputenc} 
\usepackage[T1]{fontenc}    
\usepackage{hyperref}       
\usepackage{caption}        
\usepackage{url}            
\usepackage{nicefrac}       
\usepackage{microtype}      
\usepackage{enumitem}       
\usepackage{booktabs}       
\usepackage{graphicx}
\usepackage{amsmath}
\usepackage{amsfonts}
\usepackage{amssymb}
\usepackage{amsthm}
\usepackage{soul}

\newcommand{\eg}{\textit{e}.\textit{g}., }

\newcommand{\iid}{i.i.d. }

\newcommand{\algebra}{\mathcal{A}}

\newcommand{\abs}[1]{\left|#1\right|}

\newcommand{\oneto}[1]{{1,\ldots,#1}}

\theoremstyle{definition}
\newtheorem{definition}{Definition}
\theoremstyle{plain}

\theoremstyle{remark}

\newcommand{\h}{\texttt{$h$-block} }
\newcommand{\hv}{\texttt{$hv$-block} }
\newcommand{\cv}{cross-validation}

\title{$hv$-Block Cross Validation is \emph{not} a BIBD: \\a Note on the Paper by Jeff Racine (2000)}

\author{%
  Wenjie Zheng\thanks{www.zhengwenjie.net} \\
  Independent Researcher\\
  \texttt{contact@zhengwenjie.net} \\
}

\begin{document}

\maketitle

\begin{abstract}
This note corrects a mistake in the paper \textit{consistent cross-validatory model-selection for dependent data: $hv$-block cross-validation} by \citet{racine2000consistent}.
In his paper, he implied that the therein proposed \hv cross-validation is consistent in the sense of \citet{shao1993linear}.
To get this intuition, he relied on the speculation that \hv is a balanced incomplete block design (BIBD).
This note demonstrates that this is not the case, and thus the theoretical consistency of \hv remains an open question.
In addition, I also provide a Python program counting the number of occurrences of each sample and each pair of samples.
\end{abstract}

\section{Introduction}
Cross-validation has been an important and popular tool for statistics and machine learning.
Though the \iid case is well investigated, the topic on \emph{dependent} (yet stationary) sequences is less visited.
Among these efforts, \citet{carlstein1986use}, \citet{kunsch1989jackknife}, \citet{lele1991jackknifing}, C.\ K.\ Chu (in his Ph.D. thesis in University of North Carolina), \citet{gyorfi1989nonparametric}, \citet{burman1994cross} can be considered as the pioneers in this area.
To tackle the dependence within the data, they all introduced the concept of \emph{gap}, which ``blocks'' between the training data and the test (or validation) data.
Their approaches elegantly mitigated the issue of dependence.

Following these precursors, \citet{racine2000consistent} studied the impact of the findings of \citet{shao1993linear, shao1996bootstrap} on the \h cross-validation proposed by \citet{burman1994cross}.
He discovered that \h (which uses a single validation sample in each run) is \emph{not} consistent in the sense of \citet{shao1993linear}.
That is, \h does not choose the most \emph{concise} correct model when the sample size $n \rightarrow \infty$; instead, it has an incline for larger models.
To cope with this issue, he mimicked the strategy of \citet{shao1993linear} by using $n_v:=2v+1$ validation samples each time, with $n_v / n_c \rightarrow \infty$, where $n_c$ is the size of the training set.
His strategy has since been dubbed \hv cross-validation.

\citet{racine2000consistent} is positive, both empirically and theoretically, about the consistency of \hv cross-validation.
Empirically, he found that \hv is more consistent than \h as well as the regular cross-validation when $n$ is large \citep[see][Table 3--5]{racine2000consistent}.
Theoretically, he did not ``attempt'' a direct proof; instead, he suggested that \hv is a balanced incomplete block design (BIBD) and thus could reuse the proof of \citet{shao1993linear}.
\hv has since been believed consistent, although a rigorous, documented proof is absent.

In this note, I will show you that 
\begin{center}
\hv is not a BIBD.
\end{center}
Detailed analysis, visual illustrations, and software will be provided to strengthen this argument.
The consequence of this argument is, obviously, that the theoretical consistency of \hv remains an open question and warrants further investigation.

After finding this mistake of \citet{racine2000consistent}, I also checked 64 later papers\footnote{The total number counts to 129 as of Oct. 18, 2019. The majority of papers under the radar here is those (openly) inaccessible ones. See \url{https://github.com/WenjieZ/hv-block-is-not-a-BIBD/blob/master/citation.ods} for the full list.} citing \citet{racine2000consistent}.
Among these papers, 54 fail to point out this mistake, and 10 inherit it.
In this epoch where cross-validation becomes increasingly important and \hv or similar ideas are being rejuvenated in various domains such as energy forecasting \citep{cui2016short}, medicine \citep{eisenbarth2016multivariate}, ecology \citep{roberts2017cross, valavi2018blockcv}, and financial investment \citep{de2018advances}, the erratum proposed in this note is, I believe, timely.

\section{Balanced incomplete block design}
This section presents the concept of balanced incomplete block design. 
The concept of \textbf{design} has many applications (\eg fixed and random designs of experiments in regression).
Latin squares might be the most famous example of design.
Though these concepts are intelligently intriguing, a thorough presentation of the design theory is not attempted here.
Interested readers are referred to \citet{Stinson2003combinatorial}.
Here, not to deviate from our goal, I will introduce the mere necessary.

\begin{definition}\label{def:design}
A \emph{design} is a pair $(X, \algebra)$ such that the following properties are satisfied:
\begin{enumerate}[noitemsep,nolistsep]
\item $X$ is a set of elements called \emph{points}, and
\item $\algebra$ is a collection of nonempty subsets of $X$ called \emph{blocks}.
\end{enumerate}
\end{definition}

\begin{definition}\label{def:bibd}
Let $n$, $k$, and $\lambda$ be positive integers such that $n>k\ge2$. A $(n,k,\lambda)$-\emph{balanced incomplete block design} (abbreviated to $(n,k,\lambda)$-BIBD) is a design $(X, \algebra)$ such that the following properties are satisfied:
\begin{enumerate}[noitemsep,nolistsep]
\item $\abs{X}=n$,
\item each block contains exactly $k$ points, and
\item every pair of distinct points is contained in exactly $\lambda$ blocks.
\end{enumerate} 
\end{definition}

The following is a $(7,3,1)$-BIBD:
\begin{align*}
X = & \{1,2,3,4,5,6,7\}, \\
\algebra = & [123, 145, 167, 246, 257, 347, 356],
\end{align*}
where there are 7 points, each block contains 3 points, and each pair of points occurs in 1 block.

The following is a $(7,3,2)$-BIBD:
\begin{align*}
X = & \{1,2,3,4,5,6,7\}, \\
\algebra = & [123, 145, 167, 246, 257, 347, 356, \\
& \phantom{[} 123, 147, 156, 245, 267, 346, 357],
\end{align*}
where there are 7 points, each block contains 3 points, and each pair of points occurs in 2 blocks.

Definition~\ref{def:bibd} seems to be different from the one in \citet{shao1993linear}, but in fact they are equivalent.
In his paper, he replaced the second condition of Definition~\ref{def:bibd} with the condition that each point occurs in exactly $r$ blocks.
By straightforward deduction, one can prove that $r(k-1) = \lambda(n-1)$. That is, there exists a one-to-one relation between $k$ and $r$, given $n$ and $\lambda$. 

In addition to the above equality, we have another one.
Let $b$ be the number of blocks, we can easily prove that $bk = nr$.
In the remaining of this note, I will use the overparameterized notation $(n, k, b, r, \lambda)$-BIBD.

\section{\hv cross-validation}
This section presents the \hv cross-validation proposed by \citet{racine2000consistent}.
This approach is suitable for dependent data, such as temporal, spatial, hierarchical, or phylogenetic data \citep{roberts2017cross}.

\begin{figure}
\centering
\includegraphics[width=0.5\linewidth]{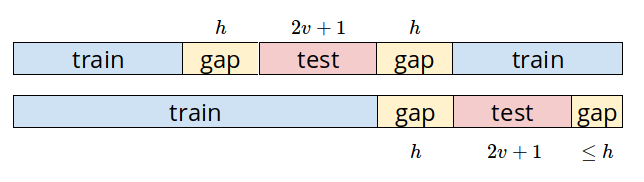}
\caption{\small Upper: the test set is far away from the boundary. Lower: the test set is near the boundary.}
\label{fig:hv-block} 
\end{figure}

The idea is to divide the data into three parts: the training set, the test set (or validation set), and the \emph{gap}.
To ease the explanation, I will use time series as an example.
Let $\{X_i\}_{i=\oneto{n}}$ be a time series, where $X_i$'s can be either random variables or random vectors.
In each validation run, we construct a contiguous test set of size $2v+1$ with the center at $X_i$.
Then, we further remove another $h$ samples (as the gaps) on both sides of the test set, and we get the training set.
When the test set is near the boundary of the whole data set and there are less than $h$ samples to remove, we just discard these remaining samples entirely (Fig.~\ref{fig:hv-block}).
Since the test set must be contiguous and contain exactly $2v+1$ samples, there are $n-2v$ legitimate test sets.

Denote $L=L(X; i, h, v)$ as the evaluation criterion (\eg empirical risk, misclassification rate) applied on the aforementioned time series $X$.
$L$ depends on the train-test split configuration.
In particular, $L(X; i, h, v)$ corresponds to the configuration using $n_v:=2v+1$ samples surrounding $X_i$ as the test set; the training set consists of the remaining samples after removing the test set and the gap.
The \hv \cv function is then given by
\[
\texttt{CV}_{hv} = \frac{1}{n-2v} \sum_{i=v+1}^{n-v} L(X; i, h, v),
\]
the average over all legitimate configurations.
Note that the starting index of $i$ should be $v+1$ instead of $v$, where  \citeauthor{racine2000consistent} made a typo in his original paper \citeyearpar{racine2000consistent}.
This typo was also observed by \citet{white2006approximate}.

The parameter $h$ controls the level of dependence between the training set and the test set.
It can either be small such that $h/n \rightarrow 0$ as required by Chu (in the thesis mentioned above) and \citet{gyorfi1989nonparametric}, or it can be large such that $h/n \equiv p$ for some $p \in (0, \tfrac{1}{2})$ as required by \citet{burman1994cross}.

As to the parameter $v$, there can be multiple choices.
If $v \equiv 0$, it becomes \h \cv{}  \citep{burman1994cross}, and \citet{racine2000consistent} indicated that it is not consistent in the sense of \citet{shao1993linear}.
A constant $v$ larger than 0 will not fix the inconsistency either.
\citet{racine2000consistent} requires that $v$ should be large such that $n_v / n_c \rightarrow \infty$, where $n_c$ is the size of the training set.

Concerning the consistency of \hv \cv, \citeauthor{racine2000consistent}'s experiments show very positive results compared to \h and regular \cv{} without gaps \citep[see][Table 3--5]{racine2000consistent}.
Also, he claimed that ``conditions required for the validity of Shao (1993) results are indeed met by the proposed \hv method'' \citep{racine2000consistent}, by which he implied that the theoretical consistency of \hv is within reach.

The condition he relied on is nothing else but the \emph{balanced incomplete block design}.
He believed that \hv is a BIBD and thus could take a free ride on \citet{shao1993linear}.

\section{\hv is not a BIBD}
In this section, I will provide three evidences, each of which single-handedly proves that \hv is \st{guilty} \emph{not} a BIBD.

\subsubsection*{Evidence 1: the two equalities of BIBD are not satisfied.}
Remember that for every $(n, k, b, r, \lambda)$-BIBD, the following two equations should hold:
\begin{align*}
r(k-1) &= \lambda(n-1), \\
bk     &= nr,
\end{align*}
which means that given the value of $(n,k,b)$, the value of $(r,\lambda)$ is determined.
If \hv is ever a BIBD, it must be a $(n, 2v+1, n-2v, x, y)$-BIBD according to the analysis in the previous section, where $x$ and $y$ are integers undetermined.

A quick calculation yields
\begin{align*}
x &= \tfrac{(n-2v)(2v+1)}{n}, \\
y &= \tfrac{2v(n-2v)(2v+1)}{n(n-1)}.
\end{align*}
As agreed, $x$ and $y$ should be integers regardless of the values of $n$ and $v$.
For instance, for $(n,v)=(10,1)$, we would expect integer values from $(x,y)$.
Nevertheless, we get $(x,y)=(\nicefrac{12}{5}, \nicefrac{8}{15})$.
This indicates that \hv is not a BIBD, at least not for all pairs of $n$ and $v$.

\subsubsection*{Evidence 2: the analytic formula for $(r,\lambda)$ protests.}
It is not difficult to explicitly calculate the occurrence of each sample and each pair of samples.
The following table gives the value of occurrence $r$ for the $i$-th sample:
\begin{center}
\begin{tabular}{cc} \toprule
$i$ & $r$ \\ \midrule
$v+1$ & $v+1$ \\ 
$v+2$ & $v+2$ \\ 
$\cdots$ & $\cdots$ \\
$2v+1$ & $2v+1$ \\
$2v+2$ & $2v+1$ \\
$\cdots$ & $\cdots$ \\
$n-2v$ & $2v+1$ \\
$\cdots$ & $\cdots$ \\
$n-v-1$ & $v+2$ \\
$n-v$ & $v+1$ \\ \bottomrule
\end{tabular}
\end{center}

We can observe that the value of $r$ first increases and then decreases.
It is a constant if and only if $v=0$, in which case it degenerates to \h \cv{} \citep{burman1994cross}.

The result for $\lambda$ is even worse.
For instance, the following matrix shows $\lambda$'s value for $(n, v)=(10, 2)$:
\[
\begin{matrix}
1 & 1 & 1 & 1 & 1 & 0 & 0 & 0 & 0 & 0 \\
1 & 2 & 2 & 2 & 2 & 1 & 0 & 0 & 0 & 0 \\
1 & 2 & 3 & 3 & 3 & 2 & 1 & 0 & 0 & 0 \\
1 & 2 & 3 & 4 & 4 & 3 & 2 & 1 & 0 & 0 \\
1 & 2 & 3 & 4 & 5 & 4 & 3 & 2 & 1 & 0 \\
0 & 1 & 2 & 3 & 4 & 5 & 4 & 3 & 2 & 1 \\
0 & 0 & 1 & 2 & 3 & 4 & 4 & 3 & 2 & 1 \\
0 & 0 & 0 & 1 & 2 & 3 & 3 & 3 & 2 & 1 \\
0 & 0 & 0 & 0 & 1 & 2 & 2 & 2 & 2 & 1 \\
0 & 0 & 0 & 0 & 0 & 1 & 1 & 1 & 1 & 1
\end{matrix}
\]
For $i \neq j$, the $(i,j)$-th entry is the occurrence for the pair $(i,j)$.
For $i = j$, the $(i,i)$-th entry is the occurrence for the single sample $i$.
Once again, these values are not constant.

\subsubsection*{Evidence 3: the software protests.}
I have developed a tool which naively counts the exact occurrence of every and every pair of samples.
It is accessible from 
\begin{center}
\url{https://github.com/WenjieZ/hv-block-is-not-a-BIBD}
\end{center}
It seems to be an overkill for the mere purpose of disproving \citet{racine2000consistent}.
However, if, one day, a weighted \hv \cv{} is attempted, this tool can be of interest.

\section{Conclusion}
This note shows that \hv \cv{} is not a BIBD, and thus the theoretical consistency of \hv remains an open question.

\subsubsection*{Acknowledgments}
I am deeply grateful to Jeff Racine, whose publication is being ``attacked'' here, for kindly encouraging me to write this note. From him as well as his manuscript, I learned what a modest, honest scientific worker should be like.

\small
\bibliographystyle{agsm}
\bibliography{hv-block}

\end{document}